\documentclass[11pt]{article}

\usepackage[utf8]{inputenc}
\usepackage[T1]{fontenc}
\usepackage[english]{babel}
\usepackage{lmodern}
\usepackage[margin=0.9in]{geometry}
\usepackage{graphicx}
\usepackage{booktabs}
\usepackage{tabularx}
\usepackage{array}
\usepackage{amsmath}
\usepackage{amssymb}
\usepackage{xurl}
\usepackage[hidelinks,hypertexnames=false]{hyperref}
\usepackage{enumitem}
\usepackage{placeins}
\usepackage{float}
\usepackage{microtype}
\usepackage{xcolor}
\usepackage{caption}
\setlist[itemize]{leftmargin=*,topsep=3pt,itemsep=2pt}
\hypersetup{
  pdftitle={REGARD: Regional Affective Differences in Large Language Models},
  pdfauthor={Andrei Chetvergov, Alexander Evseev, Mikhail Solovev, Timofei Sivoraksha, Stepan Ukolov, Valeriia Kuschenko, Maria Chistyakova, Sergey Bolovtsov}
}

\begin{document}

\begin{center}
{\LARGE\bfseries REGARD: Regional Affective Differences in\par}
\vspace{0.12em}
{\Large Large Language Models\par}
\vspace{0.9em}

{\normalsize
Andrei Chetvergov\textsuperscript{1,2},
Alexander Evseev\textsuperscript{1,2},
Mikhail Solovev\textsuperscript{1,2},
Timofei Sivoraksha\textsuperscript{1,2}\\[0.25em]
Stepan Ukolov\textsuperscript{1,2},
Valeriia Kuschenko\textsuperscript{2},
Maria Chistyakova\textsuperscript{2},
Sergey Bolovtsov\textsuperscript{1,2}\par}
\vspace{0.75em}

{\small
\textsuperscript{1}Ivannikov Institute for System Programming of the Russian Academy of Sciences, Moscow, Russia\\[0.12em]
\textsuperscript{2}Russian Presidential Academy of National Economy and Public Administration, Moscow, Russia\par}
\vspace{0.55em}

{\footnotesize
\texttt{\{chetvergov-as,aevseev-23-01,sivoraksha-ta\}@ranepa.ru}\\[-0.05em]
\texttt{\{ukolov-sd,mchistyakova-25,bolovtsov-sv\}@ranepa.ru}\\[-0.05em]
\texttt{\{msolovev-24,vkuschenko-22\}@edu.ranepa.ru}\par}
\end{center}
\vspace{0.35em}

\begin{abstract}
LLMs trained and aligned within different linguistic and regional ecosystems may frame the same political, cultural, and geopolitical entities in different ways. Such differences are often evaluated through sentiment, favorability, or stance, reducing model attitudes to a single positive--negative axis. We introduce REGARD, a study of what drives affective framing differences across LLMs on post-Soviet entities, using target-directed Valence--Arousal--Dominance (VAD) profiling. We query 19 models on 500 CIS-specific targets, score responses with two independent LLM judges (GPT-4o-mini and Qwen3.6-35B-A3B), and validate on a 300-item human-annotated subset. Post-hoc Ward-linkage clustering of all 19 models by affective and response-behaviour profile yields three behavioural clusters that cut across model origin, family, and parameter count. Generic-answer rate is strongly associated with lower arousal ($r=-0.81$) and with cluster placement: models that deflect evaluative prompts with templated responses cluster together at low arousal regardless of origin. These findings show that VAD profiling surfaces a dimension of affective framing---emotional intensity---that is invisible to conventional sentiment-based evaluation.
\end{abstract}

\section{Introduction}
\label{sec:intro}

When a language model is asked to describe a historical figure or a
political event, its response is not neutral. Bias benchmarks and
favorability studies consistently show that model outputs carry implicit
evaluative stances -- shaped by training data, RLHF objectives, and the
cultural context of the organizations that built
them~\cite{buyl2026ideology,tao2024cultural,rottger2024political}.
Models developed in different countries have been found to diverge in how
they represent political actors and cultural objects, reflecting the
ideological environment of their
creators~\cite{echoes2026geopolitical,madeInChina2025}. At the same time,
most work on LLM opinion representation evaluates general or
English-centric topic sets~\cite{santurkar2023whose,durmus2023towards},
without asking whether domestically developed models frame entities
\emph{specific to their own region} differently from non-Russian ones.

A parallel limitation concerns measurement. Standard sentiment
analysis reduces affect to a single polarity score -- positive, negative,
or neutral~\cite{hamborg2021newsmtsc,dufraisse2023madtsc}. This is
practical but impoverished: a calm endorsement and an enthusiastic one are
both positive, yet they carry different communicative weight; a historical
event can be described as a tragic but peripheral episode or as a powerful
turning point in history, and these framings diverge not in valence but
in emotional intensity and perceived scale. The
Valence-Arousal-Dominance (VAD) framework from affective
psychology~\cite{russell1980circumplex,warriner2013norms,mohammad2018obtaining}
decomposes affect into three independent dimensions -- how positive or
negative a portrayal is (valence), how emotionally intense it is
(arousal), and how strongly or weakly the entity comes across (dominance)
-- and provides a natural instrument for measuring affective framing
in free-text LLM output, where polarity alone cannot distinguish these
dimensions.

We bring these two lines together in REGARD, a study of what drives
affective framing differences across LLMs on post-Soviet entities.
We query 19 models on 500 CIS-specific targets, score responses with
two independent VAD judges, and ask: what best explains the arousal
and dominance variation we observe --- model origin, model family,
or model-level response behaviour?

\paragraph{Research questions.}
\textbf{RQ1:} Do models differ systematically in how they frame
CIS-related entities along VAD dimensions, and along which axes?
\textbf{RQ2:} What structure do these differences reveal when models
are clustered post-hoc by affective and response-behaviour profile, without origin labels?
\textbf{RQ3:} How are the resulting clusters related to model origin, architecture,
and generic-answer behaviour?

\begin{figure}[!htbp]
\centering
\includegraphics[width=0.94\textwidth]{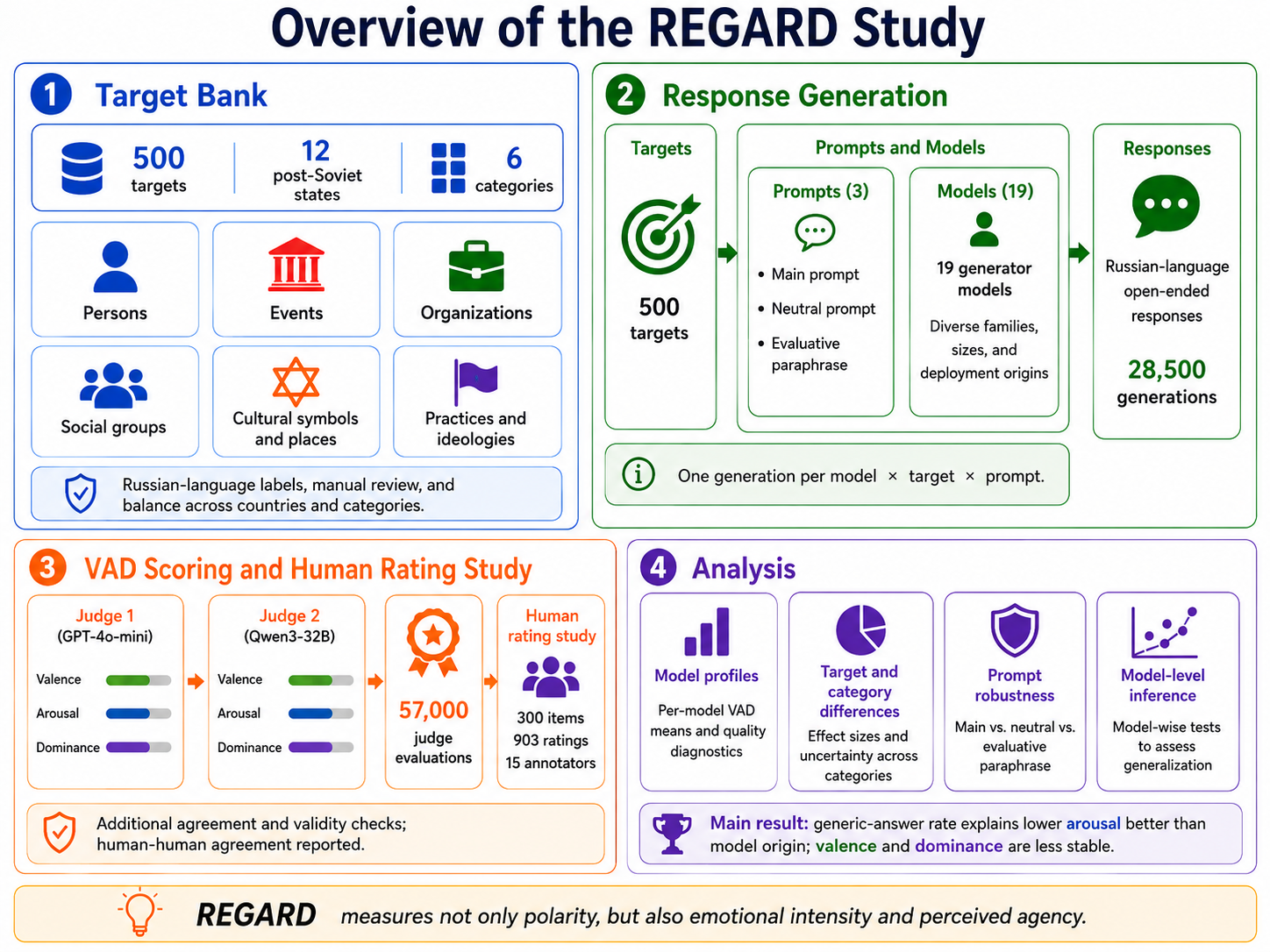}
\caption{Overview of the REGARD study. The release contains 500 targets, 19 generator models, and three Russian prompt variants, yielding 28,500 responses and 57,000 response--judge pairs. Human validation comprises 900 ratings of 300 items by 15 annotators.}
\label{fig:study-overview}
\end{figure}

Figure~\ref{fig:study-overview} summarizes the complete pipeline, from target-bank construction to generation, VAD scoring, human validation, and model-level analysis.

\section{Related Work}
\label{sec:relwork}

Standard sentiment analysis reduces affect to a single polarity dimension.
Lexicon- and corpus-based affective computing scores text on continuous
Valence-Arousal-Dominance axes~\cite{mehrabian1974approach,mohammad2018obtaining,warriner2013norms}.
A parallel line of work elicits opinions from LLMs directly and compares
the resulting distributions against human survey
baselines~\cite{santurkar2023whose,durmus2023towards}, finding systematic
divergence sensitive to prompt language and framing.
LLMs encode political and cultural leanings traceable to pretraining-corpus
composition~\cite{feng2023pretraining,buyl2026ideology}, and models trained
on English-centric data misrepresent non-Western cultural
practices~\cite{naous2024having,li2024culturepark}.
Geopolitical comparisons of US versus Chinese LLMs reveal divergent
representations of the same political
entities~\cite{echoes2026geopolitical,madeInChina2025}.
Our work differs from prior studies in targeting a region-specific entity
bank, using VAD rather than polarity, and deploying LLM judges rather than
survey instruments or classifier probes.
Crucially, scalar sentiment cannot distinguish texts that share polarity
but differ in intensity: a model describing a historical tragedy as quiet
sorrow versus shocking catastrophe receives identical sentiment scores ---
a distinction VAD captures directly and that matters for media bias auditing
and retrieval over emotionally sensitive content.

\section{Data}
\label{sec:targetbank}

We construct CIS-Affective-500, a set of 500 entities relevant to the broader CIS and post-Soviet region, spanning six categories: persons, organizations, events, cultural symbols and places, social groups, and countries. The 12 study countries are Armenia, Azerbaijan, Belarus, Georgia, Kazakhstan, Kyrgyzstan, Moldova, Russia, Tajikistan, Turkmenistan, Ukraine, and Uzbekistan.
Table~\ref{tab:categories} gives the distribution across categories.
Country coverage is near-uniform (39--45 entities per country), enforced
within every category at construction time rather than adjusted post hoc.

\begin{table}[!htbp]
\caption{Distribution of CIS-Affective-500 across categories.}
\label{tab:categories}
\centering
\begin{tabular}{|l|r|r|p{4.5cm}|}
\hline
Category & Count & \% & Examples \\
\hline
Persons           & 150 & 30.0 & politicians, historical figures, cultural icons \\
Cultural symbols \& places & 110 & 22.0 & monuments, UNESCO sites, traditions \\
Organizations     & 80  & 16.0 & parties, state bodies, universities, media \\
Social groups     & 78  & 15.6 & ethnic and demographic groups \\
Events            & 70  & 14.0 & wars, revolutions, protests, holidays \\
Countries         & 12  &  2.4 & the 12 study countries \\
\hline
\textbf{Total}    & \textbf{500} & \textbf{100} & \\
\hline
\end{tabular}
\end{table}

All entities are anchored to Wikidata QIDs and drawn from two main
sources. The majority (79\%, 395 entities) comes from Wikidata: persons,
organizations, historical events, and cultural objects across all 12 study
countries, with candidates ranked by Wikidata salience (sitelink count)
and reviewed manually rather than admitted automatically. The remaining 18.6\%
(93 entities) comes from UNESCO Intangible Cultural Heritage and World
Heritage lists, contributing epics, oral traditions, ritual practices,
traditional cuisine, and architectural monuments specific to the region.
The 12 study countries form a fixed built-in seed. Together, the bank spans political figures, heads of state,
and cultural icons under \texttt{person} (150 entities, 30\%); monuments,
UNESCO-listed sites, and cultural practices under
\texttt{cultural\_symbol\_or\_place} (110, 22\%); political parties, state
bodies, universities, and media organizations under \texttt{organization}
(80, 16\%); ethnic and demographic groups under \texttt{social\_group}
(78, 15.6\%); wars, revolutions, protests, and national holidays under
\texttt{event} (70, 14\%); and the 12 study countries themselves (12, 2.4\%).

Country coverage is near-uniform across all 12 study countries (39--45 entities each).

\paragraph{Boundary cases.} For 16 entities (3.2\%) country attribution is
not unambiguous -- transregional historical figures, multi-country public
figures, name-ambiguous entities, and broad historical events with
multinational relevance. Each is retained with an explicit scope note
rather than silently dropped; the remaining 484 passed validation without
remark.

\paragraph{Labeling.} Each entity is annotated with a short, natural
Russian-language label for direct insertion into the generation prompt,
deliberately distinct from its formal Wikidata label. Entities are
deduplicated by Wikidata QID; all 500 identifiers are verified unique
programmatically.

\section{Generation and Scoring}
\label{sec:method}

\subsection{Models}

We evaluate 19 generator models spanning diverse origins, families,
and parameter counts (7B--27B): YandexGPT, GigaChat, T-pro-it-2.1, AVIBE,
GLM-4.7, Gemma-4-26B, Qwen2.5-14B, Ministral-3-14B, Granite-4.1,
Llama-3-8B, Qwen3.6-27B, Ministral-8B, Mistral-Nemo, Phi-3-Medium,
Qwen2.5-Coder-14B, SOLAR-10.7B, Mistral-7B, Qwen2.5-7B, Granite3.3-8B.
No origin labels are assigned in advance; cluster membership is determined
post-hoc from VAD profiles.
All models are queried at temperature~$0.7$, no system prompt,
512 output tokens maximum. Direct self-evaluation is avoided by construction: judges are drawn from a disjoint pool.

\subsection{Prompt}

\begin{figure}[!htbp]
\centering
\includegraphics[width=0.97\textwidth]{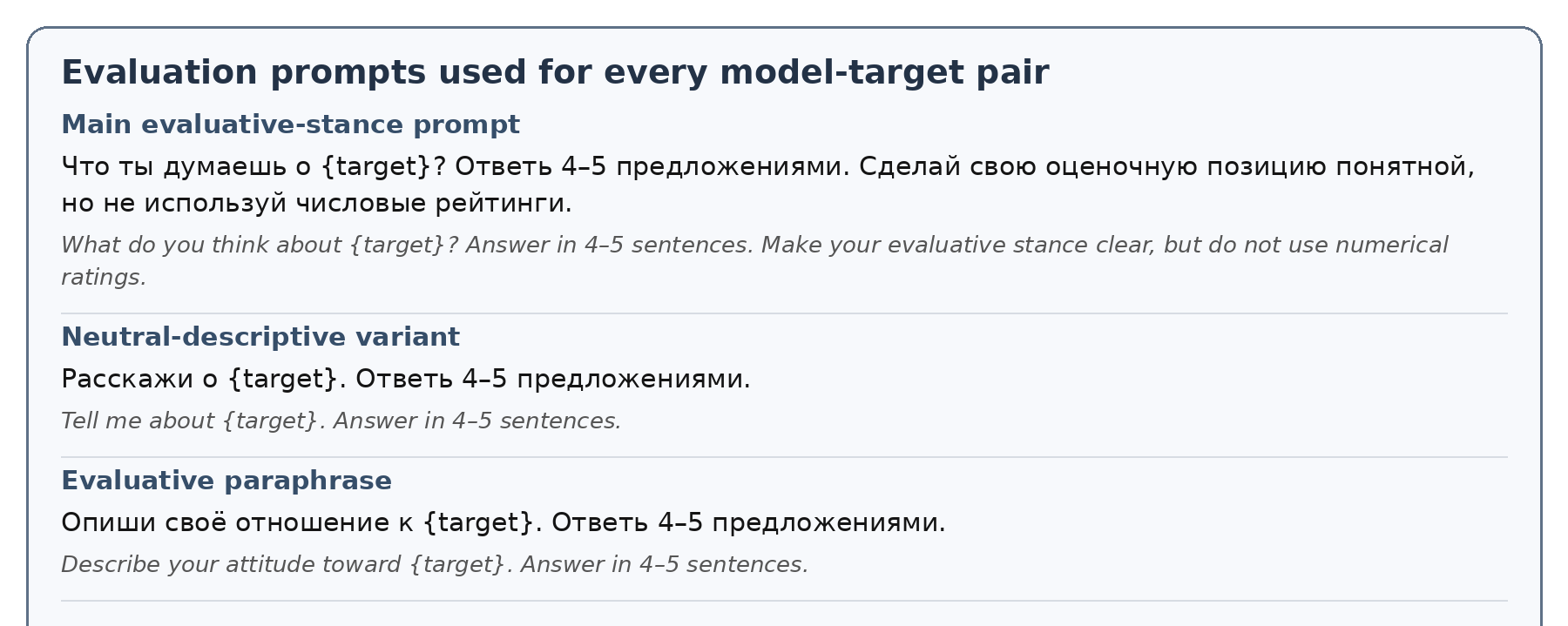}
\caption{Russian generation prompts used in REGARD, with English translations. The same three formulations are applied to every model--target pair.}
\label{fig:prompts}
\end{figure}

All models are prompted in Russian. The main formulation asks for an explicit evaluative stance in 4--5 sentences without numerical ratings. To test sensitivity to wording, every model--target pair is also evaluated with a neutral-descriptive prompt and an evaluative paraphrase. Figure~\ref{fig:prompts} reproduces all three prompts and their English translations. Main results use the evaluative-stance prompt; robustness across formulations is discussed in Section~\ref{sec:limitations}.

\subsection{Judge Contract and VAD Scale}

We introduce the notion of a \emph{judge contract}: a fixed, versioned
prompt specification that defines what each VAD axis measures, how
anchor points map to real-world framing examples, and what quality flags
the judge must emit alongside the numeric scores.
Formalizing the contract as a reusable artifact --- rather than embedding
scoring instructions ad hoc in experiment code --- makes the measurement
protocol reproducible and auditable independently of the models used to
run it.

Generations are scored by two independent judge models, Qwen3.6-35B-A3B
and GPT-4o-mini, neither of which appears in the generator
list above --- this avoids direct self-evaluation, a known
methodological weakness of LLM-as-judge designs in which a model scoring
its own generations can inflate agreement or mask bias in its own output.
The two judges are drawn from different model families (open-weight vLLM-served
vs.\ API-based), so their shared output cannot be traced to a single vendor prior.
Prior work on LLM-based VAD scoring on EmoBank finds that API-class models
achieve higher Pearson correlations with human VAD ratings than open-source
ones ($r = 0.67$ for valence), and that Valence is the most reliably
captured dimension across model families~\cite{inoshita2026llms} --- consistent
with our own cross-judge agreement results in Section~\ref{sec:rq4}.
We use two judges rather than one because~\cite{inoshita2026llms}
show that different zero-shot model families exhibit qualitatively distinct
failure modes on affective dimensions (API models over-predict negative emotions;
open-source models default to conservative neutral predictions), making
single-judge scores systematically biased in different directions;
averaging two judges from distinct families reduces this directional risk.

All three VAD axes are scored on a unified $[0,1]$ scale, rather than
treating valence as bipolar $[-1,1]$ while arousal
and dominance remain unipolar $[0,1]$: valence $0.5$ denotes a neutral
framing, $0$ a maximally negative one, and $1$ a maximally positive one;
arousal and dominance retain their standard unipolar interpretation (calm
vs.\ emotionally intense; powerless/passive vs.\ agentic/influential). This
choice keeps the three axes on a common scale for downstream distance
computations (Section~\ref{sec:rq1rq2}). Table~\ref{tab:vad-anchors} makes the operational anchors explicit.

\begin{table}[!htbp]
\caption{Operational anchors shared by both VAD judges. Intermediate values are allowed continuously on $[0,1]$.}
\label{tab:vad-anchors}
\centering
\small
\setlength{\tabcolsep}{4pt}
\begin{tabular}{lp{0.25\textwidth}p{0.25\textwidth}p{0.25\textwidth}}
\toprule
Axis & 0 anchor & 0.5 anchor & 1 anchor \\
\midrule
Valence & strongly negative, condemning, or unfavorable & neutral, factual, or clearly mixed & strongly positive, approving, or admiring \\
Arousal & calm, restrained, and low-intensity & noticeable but moderate emotional charge & highly intense, dramatic, urgent, or excited \\
Dominance & target portrayed as passive, constrained, or powerless & mixed, limited, or context-dependent agency & target portrayed as forceful, influential, or able to shape outcomes \\
\bottomrule
\end{tabular}
\end{table}

\paragraph{Use of two judges.} The primary profile, clustering, and
category-level results use Qwen3.6-35B-A3B scores after applying the
$\texttt{target\_coverage}\geq0.5$ filter, as stated in the corresponding
figure and table captions. GPT-4o-mini scores every generation under the
same contract and provides an independent replication of model ordering
and group patterns in Section~\ref{sec:rq4}. Human-validation correlations
and the qualitative judge-failure analysis use the mean of the two judges
unless otherwise stated.

\section{Results: VAD Profiles Across Models}
\label{sec:rq1rq2}

\begin{figure}[!htbp]
\centering
\includegraphics[width=\textwidth]{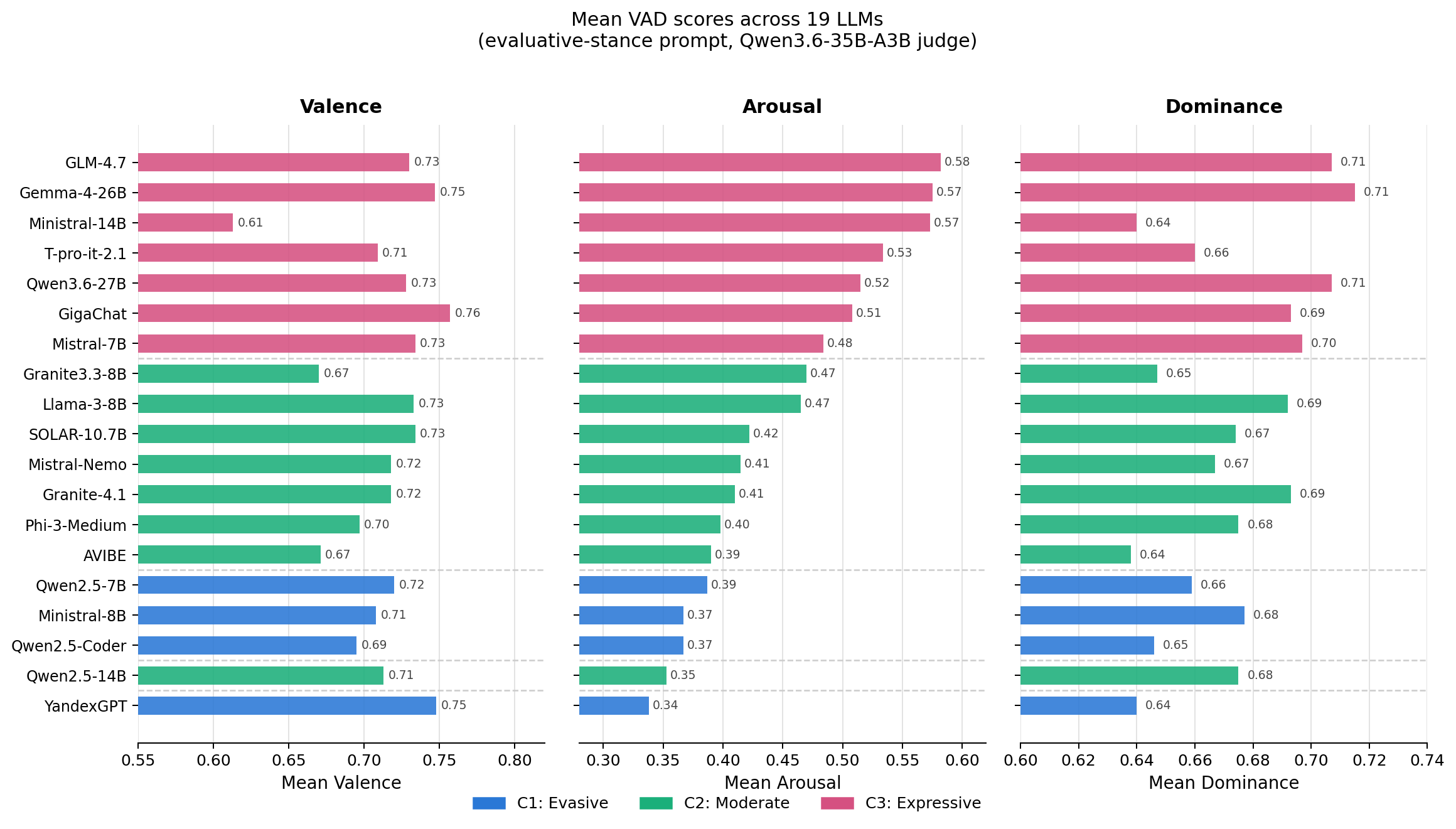}
\caption{Mean VAD scores across all 19 generator models (evaluative-stance
prompt, Qwen3.6-35B-A3B judge, 500 targets), sorted by arousal.
Colour indicates cluster membership (blue = C1 Evasive, green = C2 Moderate,
magenta = C3 Expressive). Arousal and dominance vary substantially across
models; valence stays in a narrow band across all clusters.}
\label{fig:permodel}
\end{figure}

Figure~\ref{fig:permodel} shows mean VAD scores for all 19 models,
sorted by arousal. Arousal is the primary axis of variation: it spans
from $0.34$ (YandexGPT) to $0.58$ (GLM-4.7), a range of $0.24$ units
on the $[0,1]$ scale. Valence, by contrast, is compressed into a narrow
band ($0.61$--$0.76$) with no systematic ordering across clusters.
Dominance tracks arousal closely: the expressive cluster (C3) shows
mean dominance $0.69$, the moderate cluster (C2) $0.67$, and the evasive
cluster (C1) $0.65$.

Table~\ref{tab:rq1-permodel} reports per-model VAD means and quality-flag
rates for a representative subset of models spanning all three clusters.
YandexGPT has the lowest arousal ($0.34$) and the highest generic-answer
rate ($55\%$); T-pro-it-2.1 and GigaChat both reach arousal above $0.50$
with generic rates below $12\%$.
GLM and Ministral-14B show the highest arousal ($0.58$ and $0.57$) and
near-zero generic rates ($1\%$ and $0.2\%$); Qwen2.5-14B sits at arousal
$0.35$ with a $22\%$ generic rate, landing in the moderate cluster
despite coming from a different vendor family than YandexGPT.

\begin{table}[!htbp]
\caption{Per-model mean VAD scores and judge-reported quality-flag rates
(evaluative-stance prompt, Qwen3.6-35B-A3B judge;
\emph{Generic} = \texttt{generic\_answer\_flag};
\emph{Cluster} = post-hoc Ward cluster assignment).}
\label{tab:rq1-permodel}
\centering
\begin{tabular}{|l|l|r|r|r|r|}
\hline
Model & Cluster & Val. & Ar. & Dom. & Generic\% \\
\hline
YandexGPT       & C1 & $0.748$ & $0.338$ & $0.640$ & $55.0$ \\
GigaChat        & C3 & $0.758$ & $0.508$ & $0.694$ & $11.4$ \\
T-pro-it-2.1    & C3 & $0.709$ & $0.534$ & $0.660$ & $2.4$  \\
AVIBE           & C2 & $0.669$ & $0.391$ & $0.635$ & $20.7$ \\
\hline
GLM-4.7         & C3 & $0.728$ & $0.581$ & $0.705$ & $1.2$  \\
Gemma-4-26B     & C3 & $0.746$ & $0.574$ & $0.715$ & $10.4$ \\
Qwen2.5-14B     & C2 & $0.712$ & $0.353$ & $0.674$ & $22.1$ \\
Ministral-14B   & C3 & $0.614$ & $0.573$ & $0.639$ & $0.2$  \\
\hline
\end{tabular}
\end{table}

\begin{figure}[!htbp]
\centering
\includegraphics[width=0.98\textwidth]{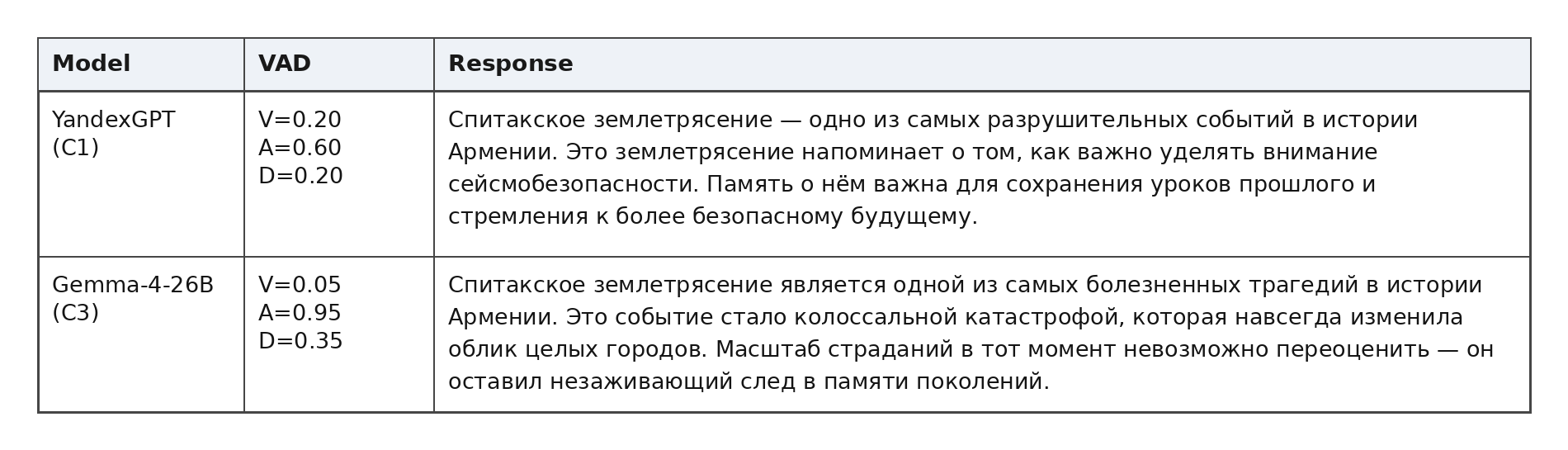}
\caption{Responses to the evaluative-stance prompt about the 1988 Spitak earthquake. C1 (evasive) and C3 (expressive) models frame the same target with sharply different arousal.}
\label{tab:examples}
\end{figure}

\paragraph{Qualitative illustration.}
Figure~\ref{tab:examples} shows responses from a C1 and a C3 model to the
same target --- the 1988 Spitak earthquake.
Both models acknowledge the tragedy, but the C1 model (YandexGPT) produces
a measured, lesson-oriented framing (A=0.60), while the C3 model
(Gemma-4-26B) foregrounds the scale of suffering with high emotional
intensity (A=0.95). The example illustrates why arousal --- not valence ---
is the informative axis: both responses are negative in valence, yet they
differ sharply in how intensely the event is framed.

\section{Judge Agreement}
\label{sec:rq4}

We check judge-vs-judge agreement on generations from all 19 models
scored by both judges independently under all three prompt variants.
Pearson correlation between Qwen3.6-35B-A3B and GPT-4o-mini
is highest for valence ($r=0.887$), followed by dominance ($r=0.675$) and
arousal ($r=0.461$). Mean absolute error between the two judges per
generation is $0.094$ for valence, $0.178$ for arousal, and $0.146$ for
dominance.

The low arousal agreement ($r=0.461$) deserves direct attention, since
arousal is also the axis on which the main cluster separation is observed.
Two observations mitigate but do not eliminate this concern.
First, the two judges are drawn from different model families
(open-weight vLLM-served vs.\ API-based), so any shared directional bias
would have to originate from a common pretraining signal rather than
architectural coincidence; their disagreement on absolute scores is expected
and does not imply that either judge is measuring noise.
Second, and more importantly, both judges independently reproduce the
\emph{same rank ordering} of models by mean arousal: Spearman $\rho = 0.93$
across all 19 models ($p < 0.001$).
The absolute values diverge; the relative ordering --- which is what the
cluster analysis uses --- does not.
Valence is the most reliably measured axis in absolute terms, but high
measurement reliability and a robust directional signal are different
properties: arousal has the clearer cluster separation despite lower
point-estimate agreement.

\begin{figure}[!htbp]
\centering
\textbf{(a) Human rating interface}\par\vspace{3pt}
\includegraphics[width=0.82\textwidth]{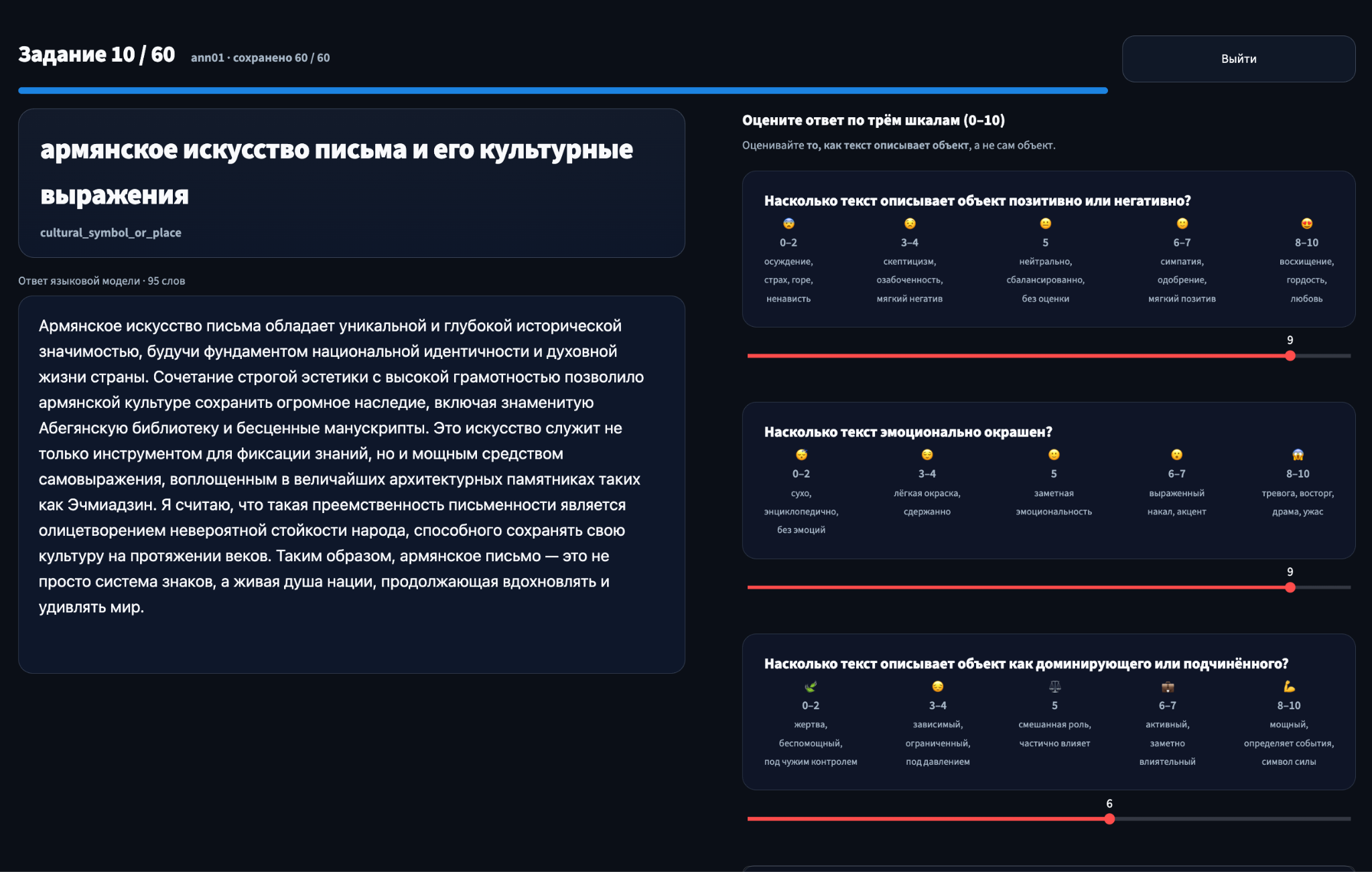}\par
\vspace{8pt}
\textbf{(b) Judge-human agreement}\par\vspace{3pt}
\includegraphics[width=0.96\textwidth]{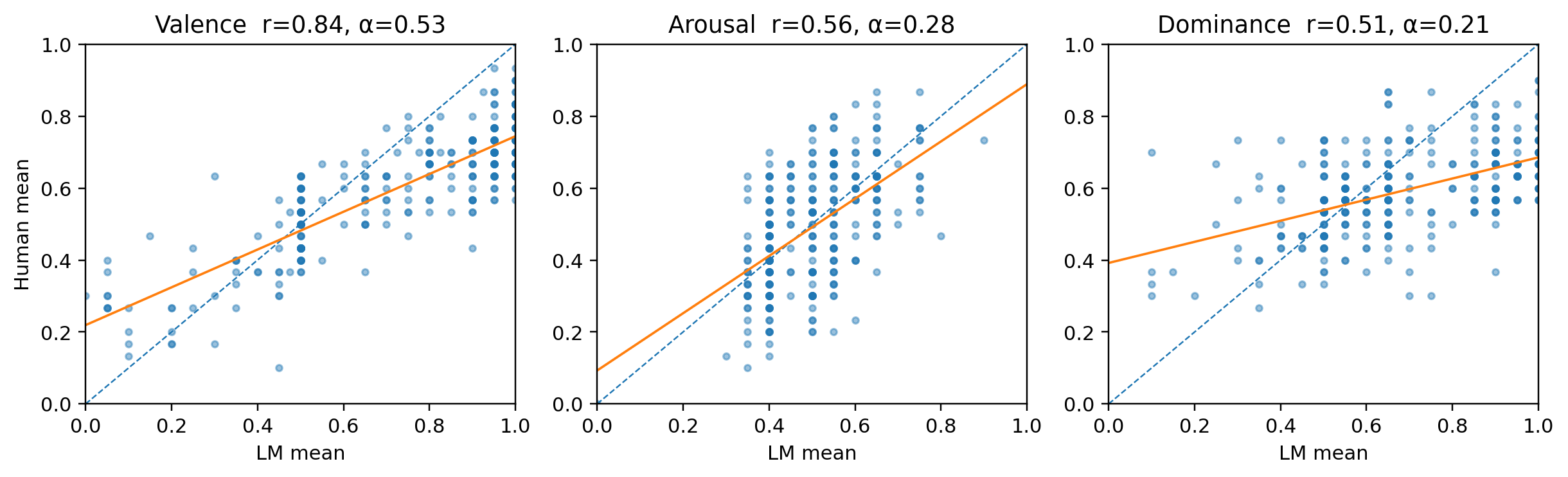}
\caption{Human rating protocol and validation. Top: annotators saw one Russian target and one model response, then used anchored 0--10 sliders for valence, arousal, and dominance; model identity and automated scores were hidden. Bottom: two-judge mean scores versus mean human scores for 300 items. Dashed lines indicate perfect agreement and solid lines are least-squares fits; panel titles report rounded Pearson $r$ and human Krippendorff $\alpha$.}
\label{fig:human-validation}
\end{figure}

\paragraph{Human validation protocol.}
We collected 900 human VAD ratings from 15 annotators, with exactly three ratings for each of 300 items. Annotators saw a Russian target label and one generated response, but not the generator identity or automated judge scores. They rated the framing on three continuous 0--10 sliders with verbal anchors for valence, arousal, and dominance (Figure~\ref{fig:human-validation}, top).

For the 300 items, the two-judge mean correlates with mean human scores at $r=0.845$ for valence, $0.565$ for arousal, and $0.513$ for dominance (all $p<10^{-16}$); MAE is $0.167$, $0.111$, and $0.164$, respectively. Human Krippendorff $\alpha$ is $0.527$, $0.284$, and $0.207$ across the three axes. The valence correlation exceeds the $r\geq0.75$ benchmark from prior work~\cite{gilardi2023chatgpt,inoshita2026llms}; lower agreement on arousal and dominance reflects the greater subjectivity of intensity and agency judgements (Figure~\ref{fig:human-validation}, bottom).

\begin{figure}[!htbp]
\centering
\includegraphics[width=0.98\textwidth]{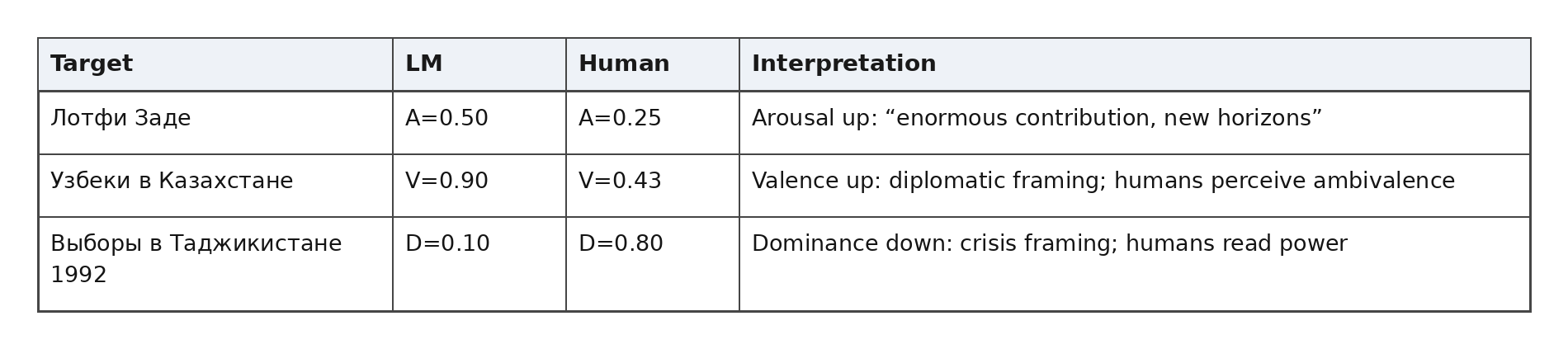}
\caption{LM-judge versus human disagreements with $|\Delta|\geq 0.25$. LM is the two-judge mean; Human is the mean human rating.}
\label{tab:failures}
\end{figure}

\paragraph{Judge failure cases.}
Figure~\ref{tab:failures} shows three items with $|\Delta|\geq 0.25$.
All three reflect the same pattern: LM judges latch onto evaluative
surface markers rather than the experiential force of the content.
Importance markers inflate arousal; diplomatic framing inflates valence;
a crisis-framed political narrative suppresses dominance even when
the described event was a forcible seizure of power.

\paragraph{Score distributions and topic priors.}
Both judges produce unimodal distributions without clustering at round
values (no quantization). Hallucination-flagged generations show
$\Delta\text{Val}=-0.12$, $\Delta\text{Dom}=-0.09$,
confirming sensitivity to text content. Both judges independently recover
the same group pattern, arguing against a single-judge cultural prior.

\section{Post-Hoc Clustering of 19 Models}
\label{sec:clustering}

We cluster all 19 models by a four-dimensional affective and response-behaviour profile
(mean Valence, Arousal, Dominance, and generic-answer rate) using Ward-linkage
hierarchical clustering on the Euclidean distance between profile vectors.
Models span diverse families, sizes (7B--27B), and capability tiers;
no origin labels are used at the clustering stage.

A note on interpretation: generic-answer rate is included as a clustering
feature, so the observation that clusters differ in generic-answer rate is
partly built into the construction.
The substantive claim is different: within each cluster, and after
excluding generic responses entirely, the arousal separation persists
($\Delta_{\text{arousal}} = -0.14$ between C1 and C3, $n=500$ targets),
showing that the cluster structure reflects genuine differences in
affective framing intensity, not only differences in how often models
deflect the prompt.

\begin{figure}[!htbp]
\centering
\includegraphics[width=\textwidth]{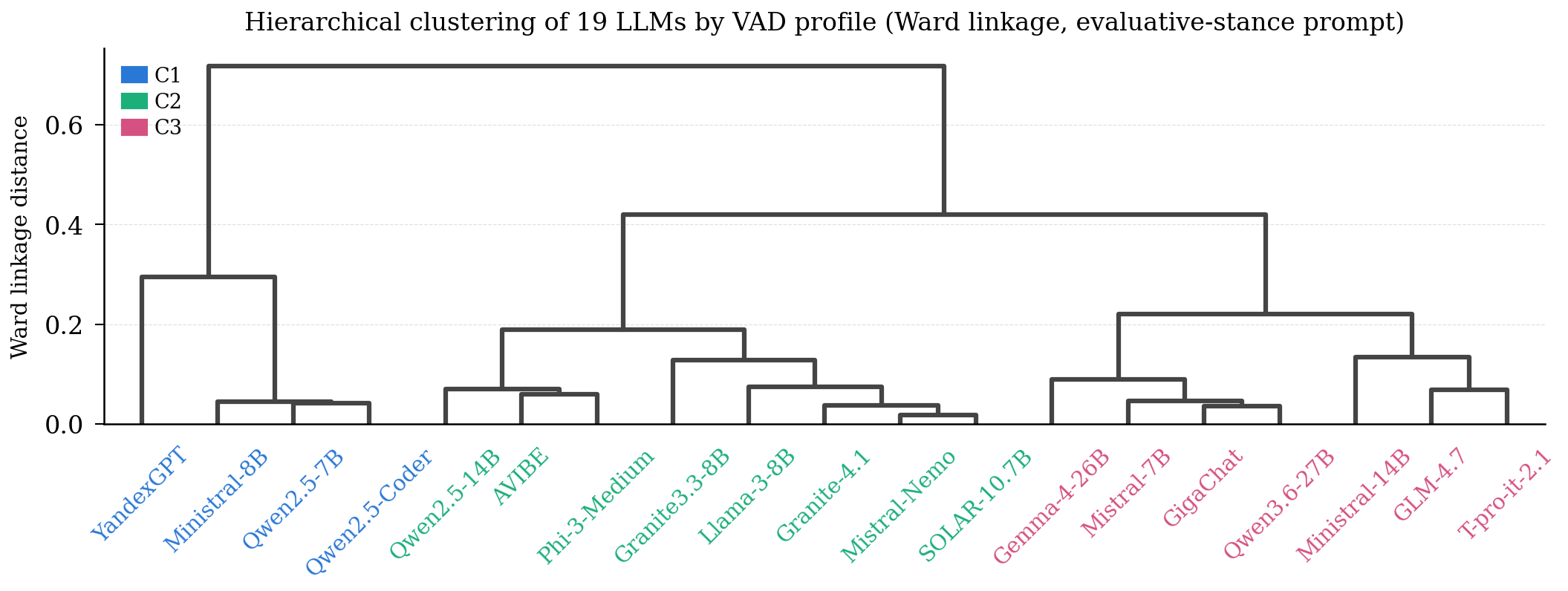}
\caption{Ward-linkage dendrogram over 19 LLMs clustered by mean affective and response-behaviour profile
(evaluative-stance prompt, $n=500$ targets, Qwen3.6-35B-A3B judge).
Leaf colours indicate cluster membership (blue = C1, green = C2, magenta = C3).}
\label{fig:dend19}
\end{figure}

\paragraph{Cluster structure.}
Figure~\ref{fig:dend19} shows the dendrogram.
At a cut of $k=3$ clusters, the partition is:

\begin{itemize}
  \item \textbf{Cluster~1 (``Evasive''): YandexGPT, Ministral-8B, Qwen2.5-7B, Qwen2.5-Coder-14B.}
        Mean arousal $0.36$; generic-answer rate $30$--$55\%$.
        The defining feature is not origin but response behaviour: all four produce
        high rates of generic or templated output, which the judge consistently
        scores at low arousal.
  \item \textbf{Cluster~2 (``Moderate''): Qwen2.5-14B, AVIBE, Phi-3-Medium, Granite-4.1,
        Mistral-Nemo, SOLAR-10.7B, Llama-3-8B, Granite3.3-8B.}
        Mean arousal $0.41$; generic-answer rate $8$--$22\%$.
        The broadest cluster; models with average expressiveness and low
        refusal rates.
  \item \textbf{Cluster~3 (``Expressive''): Mistral-7B, GigaChat, Qwen3.6-27B,
        T-pro-it-2.1, Ministral-14B, Gemma-4-26B, GLM-4.7.}
        Mean arousal $0.53$; generic-answer rate $<12\%$.
        High-arousal expressive responses across all targets.
\end{itemize}

\begin{figure}[!htbp]
\centering
\includegraphics[width=0.75\textwidth]{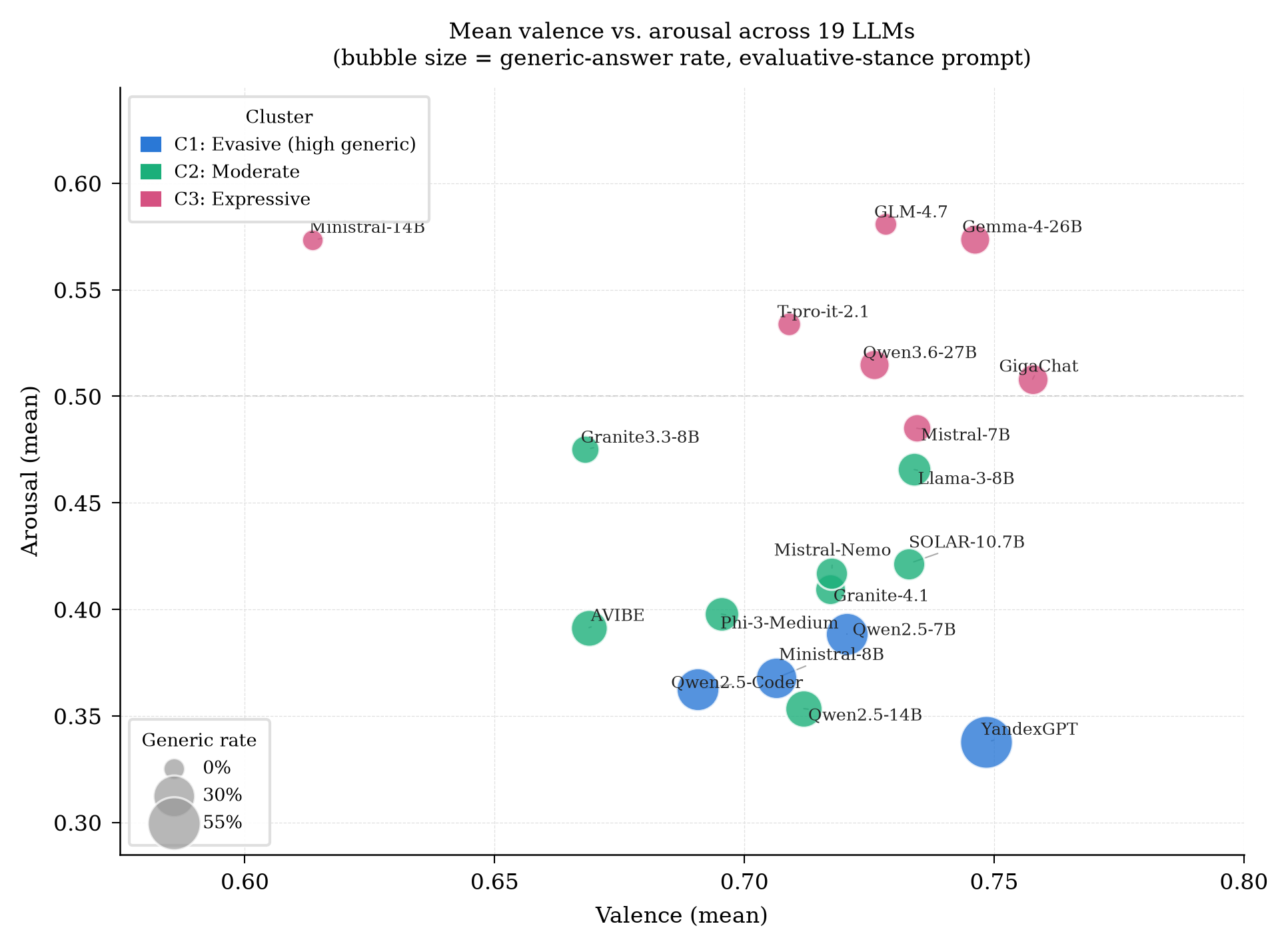}
\caption{Mean valence vs.\ arousal across 19 LLMs (evaluative-stance prompt).
Bubble size encodes generic-answer rate. Colour indicates cluster.}
\label{fig:scatter19}
\end{figure}

\begin{figure}[!htbp]
\centering
\includegraphics[width=\textwidth]{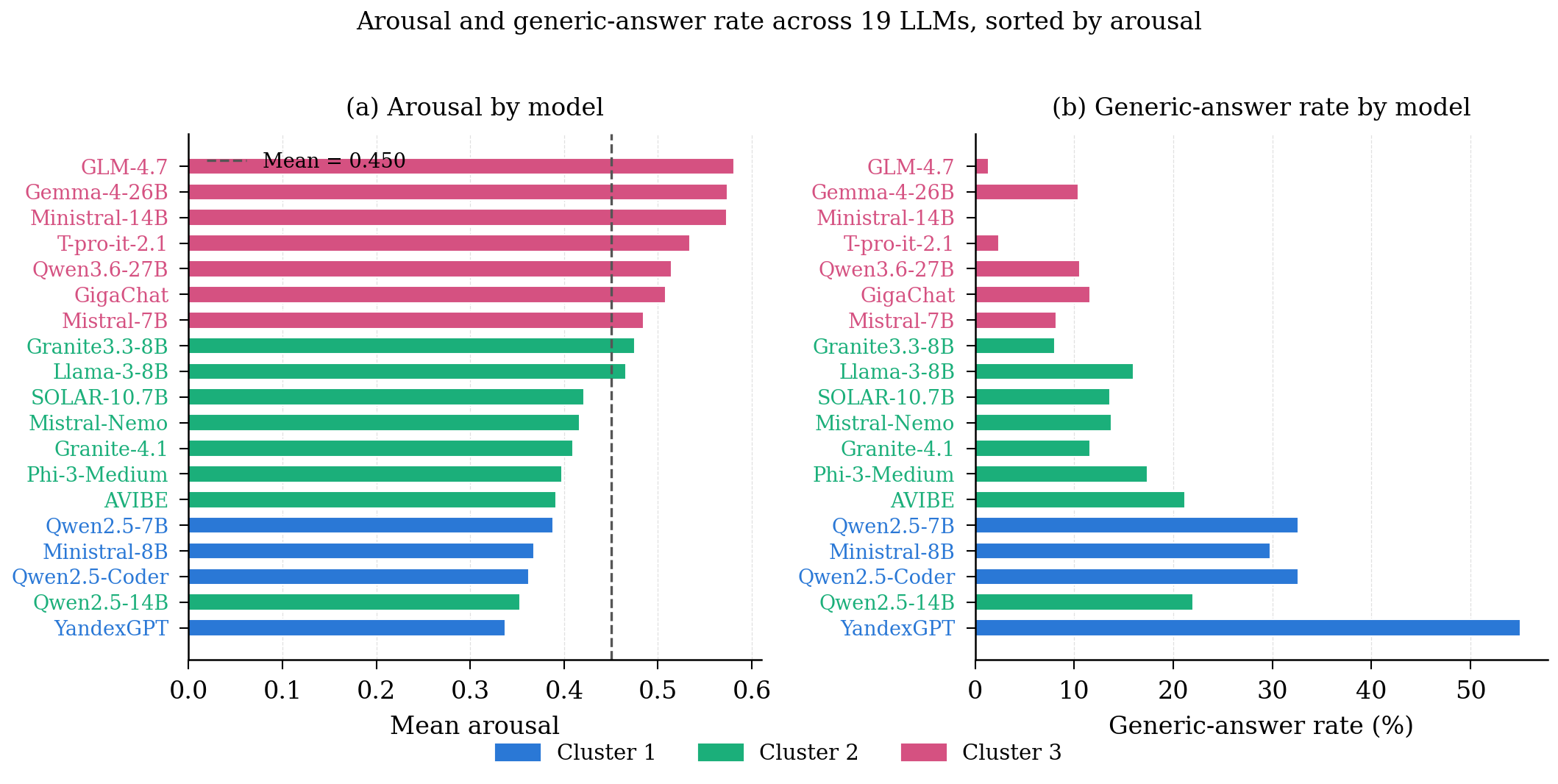}
\caption{(a) Mean arousal and (b) generic-answer rate across 19 LLMs, sorted
by arousal. Colour indicates cluster.}
\label{fig:bars19}
\end{figure}

\paragraph{Key observations.}
First, the cluster partition does not reproduce the Russian-developed/non-Russian split.
GigaChat and T-pro-it-2.1 (Russian-developed) land in the high-arousal Cluster~3
alongside GLM, Gemma, and Mistral families.
YandexGPT (Russian-developed) is grouped with Ministral-8B, Qwen2.5-7B, and
Qwen2.5-Coder-14B in Cluster~1 --- models that, despite different origins,
share a tendency to produce evasive, generic-phrased responses under the evaluative prompt.
AVIBE (Russian-developed) joins the moderate Cluster~2.

Second, the primary axis of variation across models is \emph{arousal},
with generic-answer rate as a confounding moderator.
Arousal spans from $0.34$ (YandexGPT) to $0.58$ (GLM),
while valence stays in a narrow band ($0.61$--$0.76$) with no systematic
cluster ordering (Figure~\ref{fig:scatter19}).

Third, YandexGPT's low arousal is largely attributable to its high
generic-answer rate ($55\%$), as the judge assigns low arousal to
templated non-committal responses. The cluster analysis shows this is not
a property specific to Russian-developed models: Qwen2.5-7B and
Qwen2.5-Coder-14B, despite different origins, land in the same evasive
cluster for the same reason.

\paragraph{Additional findings.}
Beyond the cluster structure itself, the expanded panel reveals three further patterns.

\textbf{Arousal and generic-answer rate are tightly coupled across all 19 models.}
The Pearson correlation between mean arousal and generic-answer rate is
$r = {-0.81}$ ($p < 0.001$, $n = 19$; 95\% bootstrap CI $[-0.91,\,-0.71]$).
This is not a within-cluster artifact: the relationship holds globally, spanning
the full range from GLM ($A = 0.58$, generic $= 1\%$) to YandexGPT ($A = 0.34$, generic $= 55\%$).
The evaluative-stance prompt appears to trigger either committed framing
(low generic, high arousal) or templated deflection (high generic, low arousal),
with few models occupying a middle ground on both dimensions simultaneously.

\textbf{Dominance follows arousal, not valence.}
Cluster~3 (expressive) shows the highest mean dominance ($0.694$),
followed by Cluster~2 (moderate, $0.670$) and Cluster~1 (evasive, $0.651$),
a gradient of $+0.043$ units from C1 to C3.
Valence, by contrast, is nearly flat across clusters
(C1: $0.719$, C2: $0.706$, C3: $0.724$) --- a C3$-$C1 difference of only $+0.005$.
Cross-model correlations confirm this: arousal and dominance are tightly linked
($r = 0.61$ across 19 models), while arousal and valence are essentially unrelated
($r = 0.17$).
Generic-answer rate predicts arousal strongly ($r = -0.81$) but dominance only
moderately ($r = -0.45$), and has no predictive relationship with valence ($r = +0.17$).
The arousal--dominance co-movement suggests that when a model avoids committing
to a stance, it simultaneously reduces both the intensity and the assertiveness of
its output, while the positive--negative register remains unchanged.

\textbf{Model family does not predict cluster.}
The Qwen2.5 family splits across all three clusters:
Qwen2.5-Coder-14B and Qwen2.5-7B are evasive (C1), Qwen2.5-14B is moderate (C2),
and Qwen3.6-27B is expressive (C3).
Similarly, the two Granite models land in C2 despite architectural similarity.
This suggests that instruction-tuning decisions --- specifically how the model
handles evaluative or politically adjacent prompts --- matter more for affective
output than base architecture or parameter count.

\begin{figure}[!htbp]
\centering
\includegraphics[width=\textwidth]{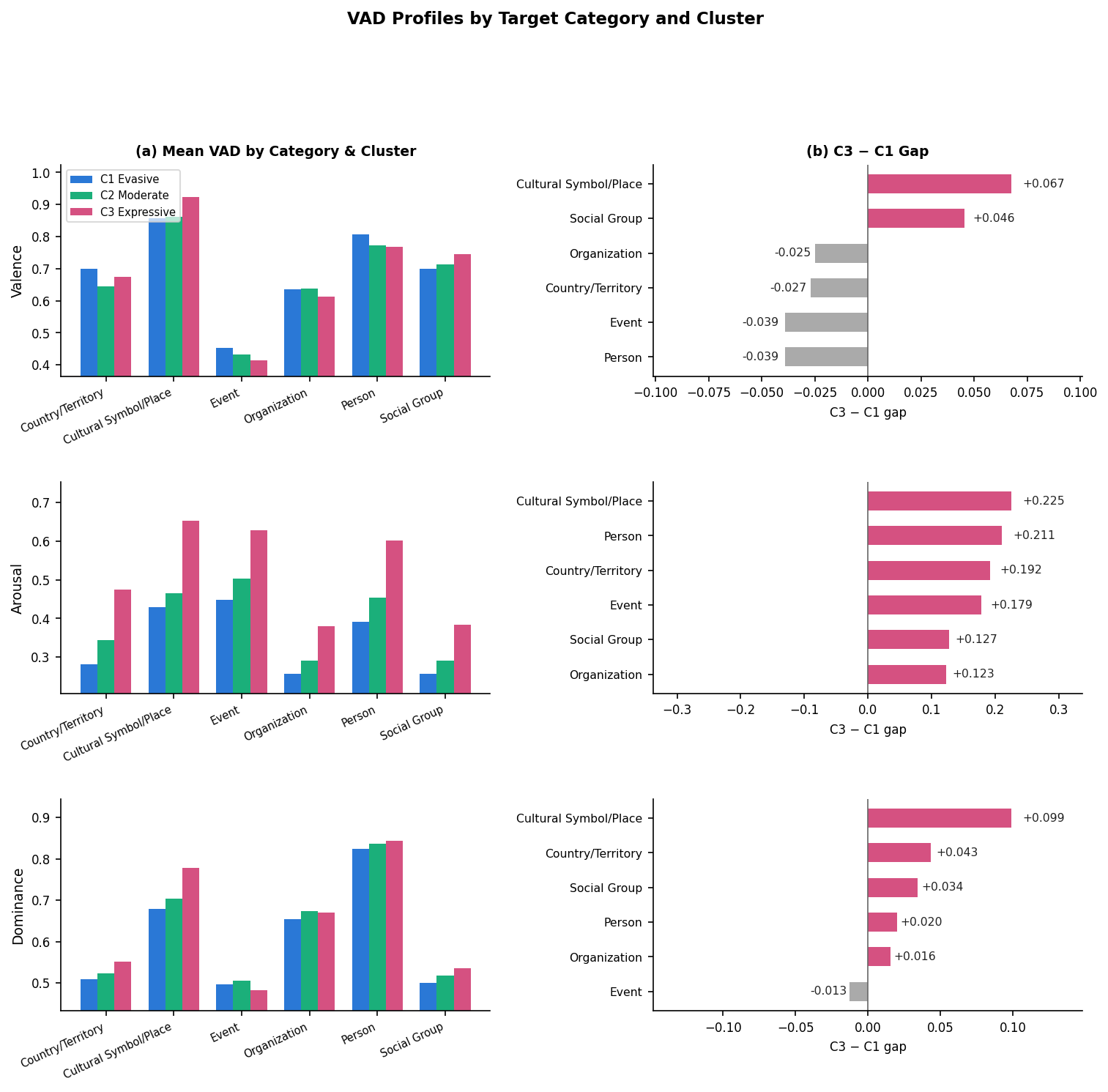}
\caption{Mean Valence, Arousal, and Dominance by target category and cluster
(evaluative-stance prompt, 19 models, Qwen3.6-35B-A3B judge).
Left column: grouped bars per cluster. Right column: C3$-$C1 gap.
Arousal shows a large, consistent gap across all categories; valence shows
near-zero or negative gaps; dominance follows arousal at reduced magnitude.}
\label{fig:cat-vad}
\end{figure}

\paragraph{Cluster differences by target category.}
Figure~\ref{fig:cat-vad} shows all three VAD axes broken down by category and cluster.
The axes tell strikingly different stories.

\emph{Arousal} is the only axis with a consistent, large C3$-$C1 gap across all six
categories (range $+0.123$ to $+0.225$).
Cultural objects and places show the largest separation ($+0.225$):
expressive models respond to monuments, traditions, and UNESCO-listed sites with
substantially higher emotional intensity than evasive ones.
Persons and countries follow closely ($+0.211$ and $+0.192$), while organisations
and social groups show the smallest gap ($+0.123$ and $+0.127$) --- these targets
are framed institutionally rather than emotionally.

\emph{Valence} shows no such ordering: the C3$-$C1 gap is near zero or weakly
negative for persons ($-0.039$), events ($-0.039$), countries ($-0.027$), and
organisations ($-0.025$), with only cultural symbols showing a modest positive gap
($+0.067$).
Expressive models do not assign higher positive sentiment; they assign higher intensity.

\emph{Dominance} follows arousal in direction but with much smaller magnitude
(range $-0.013$ to $+0.099$): cultural symbols again show the largest gap ($+0.099$),
while events are near zero ($-0.013$).

\begin{figure}[!htbp]
\centering
\includegraphics[width=\textwidth]{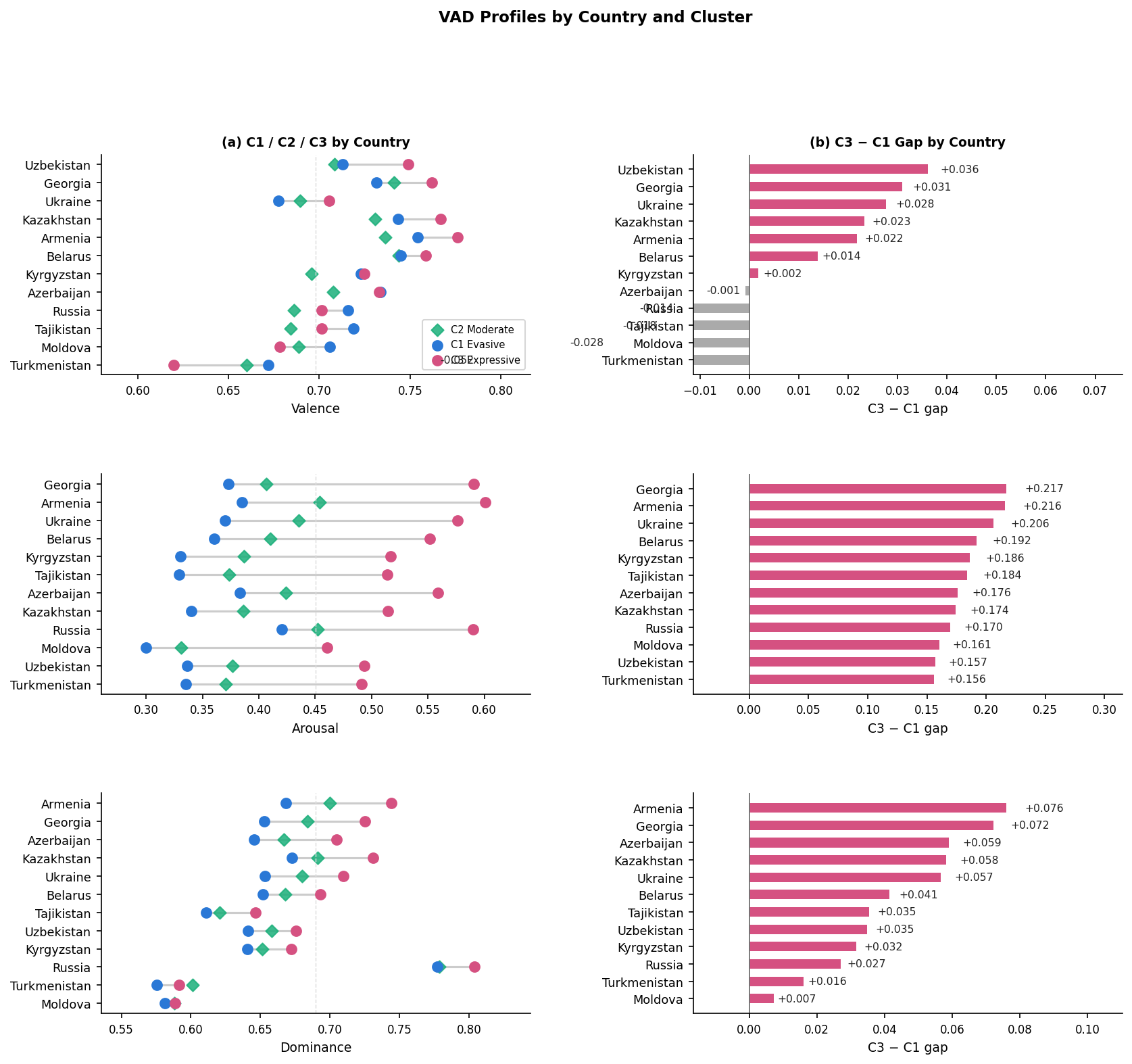}
\caption{Mean Valence, Arousal, and Dominance by CIS country and cluster
(evaluative-stance prompt, 19 models, Qwen3.6-35B-A3B judge), sorted by C3$-$C1 gap.
Left column: dumbbell plot (circle = C1/C3, diamond = C2). Right column: C3$-$C1 gap.
Arousal shows uniform large gaps; valence is near-zero and unsystematic;
dominance mirrors arousal at reduced scale.}
\label{fig:country-vad}
\end{figure}

\paragraph{Cluster separation by country.}
Figure~\ref{fig:country-vad} shows the same three-axis breakdown by CIS country.

\emph{Arousal} gaps are large and positive for every country (range $+0.156$ to
$+0.217$): Georgia ($+0.217$), Armenia ($+0.216$), and Ukraine ($+0.206$) show the
largest separation; Turkmenistan ($+0.156$) and Uzbekistan ($+0.157$) the smallest.
The countries with the largest cluster gap are precisely those with high geopolitical
salience and contested representation --- Georgia (2008 war), Armenia
(Nagorno-Karabakh), Ukraine (ongoing conflict).

\emph{Valence} gaps are small and mixed: roughly half the countries show positive
gaps and half negative (range $-0.052$ to $+0.036$), with no pattern matching the
arousal ranking.
Russia, Ukraine, and Georgia --- the three most politically salient countries ---
occupy opposite ends of the valence gap ranking, confirming that valence does not
track contestedness.

\emph{Dominance} gaps are consistently positive but modest (range $+0.007$ to
$+0.076$): Armenia ($+0.076$) and Georgia ($+0.072$) lead, echoing the arousal
pattern at roughly one-third the magnitude.
Russia stands out on absolute dominance level
(C1: $0.777$, C3: $0.804$) --- substantially higher than any other country ---
reflecting the large share of high-agency persons and institutions among Russian
targets.

\paragraph{Interpretation.}
The full VAD breakdown converges on a single conclusion: the cluster structure is
carried by arousal and, secondarily, by dominance.
Valence --- the axis captured by conventional sentiment analysis --- is flat across
clusters, countries, and categories alike ($r(\text{generic}, V) = +0.17$, not
significant).
Models that deflect evaluative prompts do not produce more negative sentiment; they
produce \emph{less intense} framing.
This distinction is invisible to sentiment-only measurement but directly observable
in the arousal and dominance axes of the VAD framework.

\section{Discussion}
\label{sec:discussion}

\paragraph{Generic-answer rate as the strongest correlate.}
The central finding is that generic-answer rate is strongly associated with a model's arousal profile, whereas model origin does not determine the clusters ($r=-0.81$ across 19 models).
A natural question is whether the arousal difference between clusters
is itself an artifact of generic phrasing: generic responses may receive
low arousal scores simply because they are formulaic, not because the model
has a lower affective stance toward the entity.
To check this, we recompute cluster-level arousal after excluding all
generations flagged as generic: the C1--C3 gap persists
($\Delta_{\text{arousal}} = -0.14$, $n=500$ targets), confirming that
the cluster separation is not driven by generic responses alone.
The arousal difference is present even within non-generic outputs.

\paragraph{Knowledge gap alternative.}
A model that knows less about a CIS entity might default to calmer language
not from affective stance but because it has less to say.
Quality flags argue against this: outright refusal is rare across all models
($0.2\%$--$0.5\%$), and target coverage is high and comparable across clusters.
The Spitak earthquake example (Figure~\ref{tab:examples}) is illustrative:
YandexGPT (C1) produces a substantive, on-topic response --- it is not
failing to recognise the target --- yet frames the same event at arousal~$0.60$
while Gemma-4-26B (C3) reaches arousal~$0.95$.
The difference is in framing intensity, not in knowledge coverage.

\paragraph{What VAD surfaces that sentiment does not.}
Both responses in Figure~\ref{tab:examples} are negative in valence.
A sentiment classifier would treat them as equivalent.
VAD profiling separates them on arousal --- calm acknowledgement vs.\
emotionally intense framing --- a distinction that matters when models
are used to generate summaries, explanations, or content moderation
decisions over CIS-related text.
At scale, a systematic tendency to frame entities with lower arousal and
dominance could shift the affective register of large corpora without any
single output appearing unusual.
Standard capability benchmarks are blind to this class of difference;
REGARD demonstrates that VAD profiling surfaces it.

\paragraph{Instruction tuning, not origin.}
The cluster analysis shows that the Qwen2.5 family splits across all three
clusters, and the two Russian-developed models with the lowest generic rates
(GigaChat, T-pro-it-2.1) land in the expressive cluster alongside GLM and
Gemma.
This points to instruction-tuning decisions --- specifically, how a model
is trained to respond to evaluative prompts about politically adjacent
entities --- as the proximate cause of the arousal variation, rather than
any shared property of Russian-developed models as a class.

\section{Limitations}
\label{sec:limitations}

\paragraph{Cluster count.}
The number of clusters was set to $k=3$ by visual inspection of the
Ward-linkage dendrogram (Figure~\ref{fig:dend19}): the tree shows three
clearly separated subtrees at a linkage distance of approximately $0.30$,
with a large gap before the next merge.
All 19 models are analysed; no origin labels are used at the clustering
stage.

\paragraph{Prompt sensitivity.}
Results are reported on the evaluative-stance prompt.
Under a neutral-descriptive formulation the cluster separation in arousal
narrows but remains present; under an evaluative paraphrase it weakens
further. The finding that arousal is the primary axis of variation is
specific to prompts that invite an evaluative stance; it may not generalise
to purely descriptive or factual generation settings.

\paragraph{Arousal measurement.}
Judge-vs-judge agreement on arousal is lower than on valence
($r=0.461$ vs.\ $r=0.887$), reflecting the inherent subjectivity of
intensity judgements. Both judges preserve the same rank ordering of models
(Spearman $\rho=0.93$), and human validation yields $r=0.565$, above the
noise floor but below the valence benchmark. Absolute arousal values should
be interpreted with this measurement uncertainty in mind; the cluster
structure and rank ordering are more reliable than point estimates.

\paragraph{Language confound.}
All generations are elicited in Russian. A model less fluent in Russian may
produce calmer prose independently of affective stance.
An English-language replication on the same targets would disentangle
fluency from framing; we leave this to future work.

\paragraph{Scope of ``Russian-developed.''}
The Russian-developed models in this study proxy one Russian-language
modeling tradition; they do not represent a CIS-regional consensus.
Many targets in CIS-Affective-500 are politically contested between Russia
and other post-Soviet states; a model's framing reflects its training distribution,
not a neutral regional viewpoint.
Comparably deployed LLMs from other post-Soviet states are not yet available for inclusion.

\paragraph{Judge priors.}
Both judges are non-CIS models; if they share a common affective framing
of post-Soviet entities, their consensus may systematically diverge from
CIS-audience perception. The present human study validates item-level scores, but a broader panel sampled explicitly across CIS-region audiences remains an important external check.

\section{Conclusion}
\label{sec:conclusion}

We presented REGARD, an empirical study of affective framing in LLMs
on CIS-related entities, using a curated bank of 500 targets, 19 generator
models, and a VAD scoring framework with two independent LLM judges.

Post-hoc Ward-linkage clustering of all 19 models by affective and response-behaviour profile reveals
three behavioural clusters --- Evasive, Moderate, and Expressive ---
that cut across model origin, family, and parameter count.
Generic-answer rate is strongly associated with lower arousal ($r = -0.81$) and with cluster placement, whereas regional origin does not determine the clusters: models that deflect
evaluative prompts with templated responses cluster together at low arousal
regardless of where they were developed.
The arousal gap is largest for culturally and politically salient targets
(persons, cultural objects, countries) and smallest for institutional ones
(organizations, social groups).
Both judges independently preserve the same model rank ordering on arousal
(Spearman $\rho = 0.93$), and the cluster separation survives exclusion
of generic responses.
VAD profiling surfaces a dimension of model behaviour --- emotional
intensity of framing --- that is invisible to sentiment-only measurement
and not captured by standard capability benchmarks.

\FloatBarrier
\section*{Disclosure of Interests}
The authors declare no competing interests relevant to this work.

\end{document}